\documentclass[conference]{IEEEtran}
\ifCLASSINFOpdf
   \usepackage[pdftex]{graphicx}
   \usepackage{adjustbox}
   \usepackage{multirow}
   \usepackage{diagbox}
\else
\fi
\usepackage{booktabs}
\usepackage{amsmath}
\usepackage{amsthm}
\newtheorem{mydef}{Definition}
\usepackage{algorithm}
\usepackage{algpseudocode}

\usepackage{booktabs}
\begin{document}
%
\title{Fingerprinting Concepts in Data Streams with Supervised and Unsupervised Meta-Information}




%
\author{\IEEEauthorblockN{Ben Halstead\IEEEauthorrefmark{1},
Yun Sing Koh\IEEEauthorrefmark{1},
Patricia Riddle\IEEEauthorrefmark{1}, 
Mykola Pechenizkiy\IEEEauthorrefmark{2},
Albert Bifet\IEEEauthorrefmark{3},
Russel Pears\IEEEauthorrefmark{4}}
\IEEEauthorblockA{\IEEEauthorrefmark{1}University of Auckland
Email: bhal636@aucklanduni.ac.nz,\{ykoh, pat\}@cs.auckland.ac.nz}
\IEEEauthorblockA{\IEEEauthorrefmark{2}Eindhoven University of Technology
Email: m.pechenizkiy@tue.nl}
\IEEEauthorblockA{\IEEEauthorrefmark{3}University of Waikato and LTCI, T{\'e}l{\'e}com Paris, IP-Paris 
Email: abifet@waikato.ac.nz}
\IEEEauthorblockA{\IEEEauthorrefmark{4}Auckland University of Technology
Email: russel.pears@aut.ac.nz}
}


\maketitle

\begin{abstract}

Streaming sources of data are becoming more common as the ability to collect data in real-time grows. A major concern in dealing with data streams is concept drift, a change in the distribution of data over time, for example, due to changes in environmental conditions.
Representing concepts (stationary periods featuring similar behaviour) is a key idea in adapting to concept drift.
By testing the similarity of a concept representation to a window of observations, we can detect concept drift to a new or previously seen recurring concept.
Concept representations are constructed using meta-information features, values describing aspects of concept behaviour.
We find that previously proposed concept representations rely on small numbers of meta-information features. These representations often cannot distinguish concepts, leaving systems vulnerable to concept drift.
We propose FiCSUM, a general framework to represent both supervised and unsupervised behaviours of a concept in a \textit{fingerprint}, a vector of many distinct meta-information features able to uniquely identify more concepts.
Our \textit{dynamic weighting} strategy learns which meta-information features describe concept drift in a given dataset, allowing a diverse set of meta-information features to be used at once.
FiCSUM outperforms state-of-the-art methods over a range of 11 real world and synthetic datasets in both accuracy and modeling underlying concept drift.

\end{abstract}


%
\IEEEpeerreviewmaketitle

\section{Introduction}
Data is becoming increasingly available in a streaming format, for example, the readings of a network-connected sensor or the online collection of website data.
One of the biggest problems in applying classification algorithms to streams is that the distribution of data can change as the stream progresses.
A data stream often features periods of stationarity known as concepts separated by changes in distribution known as concept drift.
Classifiers which fail to adapt to concept drift may suffer degraded performance.
Meta-information, or information about how the stream is behaving, can distinguish between concepts.
For example, an increase in error rate or a change in variance of a feature may indicate that a concept drift has occurred and adaptation is required.
In this research, we propose using a vector of meta-information features, or \textit{concept fingerprint}, to uniquely represent each concept.

Representing concepts such that they can be uniquely identified is important in three ways.
Firstly, in order to detect concept drift we need a representation which can identify change in recent observations.
Concept drift detection allows us to adapt a classifier to the new concept, increasing performance by dropping invalid or unnecessary behaviour.
A common adaption is to build a new classifier after each concept drift.
If two concepts cannot be uniquely represented, we will not detect a drift between them.
Secondly, concept representations can be stored and queried for future reuse.
In many cases, concepts reoccur over a stream. This is common, for example, in concepts related to seasonality or the business cycle.
We can adapt to a recurring concept by reusing the last classifier applied to the concept.
Reusing a classifier rather than building a new classifier allows knowledge to be transferred between segments, amplifying the performance benefit of adaption.
Lastly, by identifying recurring concepts we are able to track concept occurrences over the stream. This can be a valuable tool in contextualizing both the behaviour of the classification system and the environment the stream is taking place in.
All three benefits depend on the ability to distinguish between concepts using a given representation. For example, a concept representation made up of the variance of each feature may not be able to distinguish two concepts which only differ in the distribution of labels. A concept drift between the concepts would not be detectable, and we would not be able to specifically identify future recurrences.

Two broad categories of concept representations have been proposed, each differentiating between distinct concept behaviours.
Unsupervised approaches learn a representation of features seen in observations drawn from a concept.
Supervised representations additionally include the ground truth label or label assigned by a trained classifier.
Meta-information features are commonly used in both cases to represent behaviour in an efficient way.
Different meta-information features can uniquely identify different concepts. For example, unsupervised approaches may not be able to discriminate between concepts where only the labelling function has drifted while supervised methods may not differentiate concepts featuring similar labelling functions.
We hypothesize that individually, current concept representations do not uniquely identify enough concepts to work in general over a range of datasets. We show that this leads to failure cases when a dataset displays concepts unable to be uniquely represented.

We propose the Fingerprinting with Combined Supervised and Unsupervised Meta-Information (FiCSUM) framework to avoid these failure cases by unifying the benefits of current representations.
In contrast to existing methods, FiCSUM combines both supervised and unsupervised meta-information features into a fingerprint vector to capture many aspects of concept behaviour.
For example, consider a concept with an error rate of 0.3 and two features with mean and feature importance of 1, 0.8 and 0.05 and 0.2 respectively.
A simplified fingerprint representing this concept, the vector $[0.3, 1, 0.8, 0.05, 0.2]$, is able to capture a range of supervised and unsupervised behaviours. 

FiCSUM uses a fingerprint capturing at least 65 aspects of concept behaviour.
We introduce a novel dynamic weighting scheme based on feature selection methods to learn and adjust the influence of each meta-information feature online per dataset.
Dynamic weights allow the meta-information set to be general across datasets while also being flexible, allowing additional features to be added or removed as needed.
We show concept fingerprints are able to discriminate between concepts in a wide range of datasets where pure supervised or unsupervised approaches fail. 
We provide the following contributions:
\begin{itemize}
    \item The FiCSUM framework and implementation. We represent each concept identified in a stream using a fingerprint vector combining many meta-information features.
    \item We propose a method of querying a fingerprint using a weighted vector similarity measure, with weights learned dynamically to adjust sensitivity to each meta-information feature. This enables a general fingerprint construction to be used across many different datasets.
    \item Our evaluation shows that FiCSUM provides significantly better classification accuracy and ability to capture ground truth concepts than exclusively supervised or unsupervised methods. 
    We analyse the discrimination ability of systems across a range of datasets featuring drifts of different types.
\end{itemize}

In the next section, we define the recurring concept problem  and discuss concept drift detection and model selection. Section~\ref{Sec:Method} describes the FiCSUM framework. Section~\ref{Sec:Evaluation} analyses the performance of FiCSUM against other recurrent concept approaches as well as other frameworks. Section~\ref{Sec:PastWork} discusses related work and future directions of research.

\section{Background}\label{Sec:background}
We consider a data stream classification problem where a sequence of observations $\langle X, y \rangle$ is received over time, with  $X$ representing a $d$-dimensional feature vector and $y$ representing a single discrete class label.
We assume class labels are available with no delay, a common assumption but avoided by some methods such as L-CODE~\cite{zheng2019labelless}.
A concept drift is a period over which the joint distribution of observations changes, \textit{i.e.}, a concept drift from time $t_1$ to $t_n$ implies $p_{t_1}(X, y) \ne p_{t_n}(X, y)$.
We assume that between concept drifts there is a contiguous set of observations which shares some joint distribution.
We can consider a stream to be a sequence of such stationary segments, separated by concept drift.
When two distinct stationary segments share a joint distribution, we refer to them being recurrences of the same concept.

We propose learning a \textit{fingerprint} representation of each concept encountered in a data stream, allowing us to identify and track future recurrences.
By accurately tracking concepts, we can adapt a classifier to maximise performance on each concept.
Often the probability $p(y|X)$ differs between two concepts. 
Classification algorithms rely on the assumption that $p(y|X)$ is stationary in order to generalize knowledge between observations. 
A classifier trained on observations drawn from one concept may not generalize to observations drawn from a different concept, leading to a higher error rate.
By detecting concept drift, we can instead train a separate classifier over each stationary segment in a stream.
This forgets behaviour irrelevant or counterproductive to the new segment, allowing the new classifier to reach higher performance levels. 

Training separate classifiers on each segment can be inefficient when a segment is a recurrence of a previous concept. We could instead reuse the classifier from the last occurrence, transferring knowledge of the concept. Reusing a classifier can produce a lower error rate than building a separate classifier.
Identifying recurring concepts requires some representation of each concept to be learned and stored so it can be tested against future stream segments in a process known as \textit{model selection}.

Identifying recurring concepts can often be a desired feature in its own right, in addition to the performance benefit it brings, as it allows us to track occurrences of each concept across a stream. A concept occurrence defines the behaviour of both the classification system, and the environment the stream is working in.
Identifying a recurrence of a concept allows us to contextualize the behaviour of the stream and classifier in terms of the last concept occurrence.
For example, if the classifier associated with a concept had a high error rate for a specific class on previous occurrences, we may infer that a recurrence of that concept in the future will also have a high error rate for that class.
If we notice that a concept has previously only occurred during storms, we might infer from a recurrence that a storm is likely.

We define the ability to track a concept $C$ as the co-occurrence between $C$ and some concept representation $M$ built by the system. 
Here we assume each concept refers to a discrete distribution $p(y, X)$. We also assume that $M$ at each observation is discrete and associated with a single classifier. While $M$ could be associated with either a single classifier or an ensemble of classifiers, considering single classifiers allows for a discrete representation of system behaviour.
Under these assumptions co-occurrence between $C$ and $M$ can be calculated  using precision and recall.
Given the set of timesteps seen so far as $T$, each observation $o_t$  at $t\in T$ is drawn from some concept $c_t$ and is classified using classifier $m_t$. 
The precision, recall and F1 of some $M$ in tracking $C$ are given by the equations
$$ P_{CM} = \frac{|\{t \in T | m_t = M, c_t=C\}|}{|\{t \in T | m_t = M\}|}
$$
$$ R_{CM} = \frac{|\{t \in T | m_t = M, c_t=C\}|}{|\{t \in T | c_t = C\}|}
$$
$$
F1_{CM} = 2 \frac{R_{CM} P_{CM}}{R_{CM} + P_{CM}}.
$$

For each $C$ there is an $M$ which maximises $F1_{CM}$, or best tracks $C$.
We define the co-occurence F1 score, C-F1, for a system on dataset $DS$ as the average $F1_{CM}$ score achieved by approximating each $C \in DS$ with the $M$ which best tracks $C$. 
    \[
    \textrm{C-F1} = \sum_{C\in DS}\frac{1}{|DS|}\max_M F1_{CM}
    \]

The next section further describes the problems of concept drift detection and model selection. The FiCSUM framework is presented in Section~\ref{Sec:Method} as a solution.

\subsection{Drift Detection and Model Selection}\label{sec:dd_ms}
Given some concept representation $M$ describing a distribution $p(y, X)$, the goal of both concept drift detection and model selection is to decide whether a window of observations is drawn from the distribution represented by $M$.
In concept drift detection, $M$ represents our knowledge of the current stream segment, and we test if new observations are still drawn from this distribution.
In model selection, $M$ represents a previously seen segment, and we test if a new segment may be a recurrence. If we find that current observations are drawn from $M$, we identify that $M$ has reoccurred.
Previous approaches to concept drift detection and model selection have used different methods of representing $M$, requiring different methods of testing $M$ against a window of observations.

There are two broad methods for representing a joint distribution $p(y, X)$.
Supervised methods learn to represent $p(y|X)$ given a sequence of observations $O = \{o_i, o_{i+1}, \dots, o_{i+j}\}$ where each $o_i \in O$ is a pair $\langle X_i, y_i\rangle$ drawn from concept $C$.
Some supervised representations associate each $o_j \in O$ with a label $l_j$ assigned by a classifier trained on $\{o_i, o_{i+1}, \dots, o_{j-1}\}$.
A classifier learns a relationship between $X$ and $y$, so the distribution $p(l|X)$ approximates $p(y|X)$. 
In this case, the distribution is represented by the sequence of labeled observations, $\{L_i, L_{i+1}, \dots, L_{i+j}\}$ where each $L_i$ is the triple $\langle X_i, y_i, l_i\rangle$.
Unsupervised methods use changes in $p(X)$ to represent $p(y, X)$. 
These approaches represent a concept $C$ by the distribution of features $X$ seen in observations drawn from $C$, or the sequence $\{ X_i, X_{i+1}, \dots, X_{i+j} \}$.

Often representing the full distribution of $M$ is unnecessary for identifying concepts, and inefficient to store and test.
$M$ can instead be built from \textit{meta-information features} which approximate the distributions $p(y|X)$ or $p(X)$.
A meta-information feature describes one aspect of the behaviour of a given concept in a single value.
We define an unsupervised meta-information feature in Definition~\ref{def:unsup} and a supervised meta-information feature in Definition~\ref{def:sup} to categorize features describing $p(X)$ and $p(y|X)$ respectively.
A concept can be represented by a single meta-information feature, or a set of meta-information features to describe multiple aspects.
For example the set of Shapley values~\cite{zheng2019labelless} describing the importance of each feature can approximate $p(y|X)$ by describing the behaviour of a given classifier, or the mean and standard deviation of a feature can approximate $p(X)$.
These representations do not fully describe either distribution but may be enough to uniquely identify each concept.

\begin{mydef}\label{def:unsup}
An unsupervised meta-information feature $f(\{X_i, X_{i+1}, \dots, X_{i+j}\}) \rightarrow V$ maps a sequence of $d$-dimensional feature vectors drawn from some concept $C$ into a single real value $V$. This value describes one aspect of the behaviour of features drawn from $C$, or $p(X)$.
\end{mydef}
\begin{mydef}\label{def:sup}
A supervised meta-information feature $f(\{L_i, L_{i+1}, \dots, L_{i+j}\}) \rightarrow V$ maps a sequence of labeled observations where each observation $L_i$ contains a $d$-dimensional feature vector $X_i$, a ground truth label $y_i$ and a predicted label $l_i$, drawn from some concept $C$ into a single real value $V$.  This value describes some aspect of the distribution $p(y|X)$ for $C$, or the distribution $p(l|X) \sim p(y|X)$ of a classifier representing $C$.
\end{mydef}

Given a representation $M$, we can test the likelihood that a window of observations, $W$, was drawn from $M$. A partial representation, $P$, can be constructed to represent $W$ in the same format as $M$. The similarity, $Sim$, between $P$ and $M$ can then be calculated. A high similarity indicates that $W$ is likely to have been drawn from $M$. For example, using an error rate meta-information feature, $M$ and $P$ may be constructed as the mean number of observations from their respective windows where $y_i \ne l_i$. In this univariate case, $Sim$ could be the inverse absolute difference $ \frac{1}{|M - P|}$. A high $Sim$, when the error rates $M$ and $P$ are similar, indicates that the behaviour of the classifier on $W$ is similar to that represented by $M$. When $M$ is a set of meta-features, a vector similarity measure like cosine similarity could be used to calculate $Sim$.

A concept drift detector can use $Sim$ to detect concept drift.
The similarity between a representation of the current stream segment, $M_C$, and a recent window, $A$, measures whether recent observations are likely to have been drawn from $M_C$.
In stationary conditions $A$ produces a partial representation with high similarity to $M_C$. When a concept drift occurs, $A$ may produce a partial representation with low similarity to $M_C$. 
Periodically testing recent observations against $M_C$ produces a sequence of similarity values $[Sim_{1}, Sim_{2}, \dots, Sim_{n}]$.
Concept drift detectors such as ADWIN~\cite{bifet2007learning} or EDDM~\cite{baena2006early} can identify concept drift in this sequence as a cut point ${t_d}$ where values after $t_d$ are significantly different to values before $t_d$. 

Model selection processes can use $Sim$ to identify recurring concepts.
After a concept drift detection, a window of observations $W$ can be collected to represent some emerging concept.
For each stored concept representation $M$, we can construct a partial representation $P$ from $W$. In the supervised case, this may involve using the classifier associated with $M$ to obtain predicted labels $l_i$ for each observation in $W$.
Testing $M$ against $P$ produces similarity $Sim_{WM}$, representing the similarity of the distribution seen in $W$ to the representation $M$.
The model selection process then decides if $M$ is representative of $W$ given $Sim_{WM}$, for example using a simple threshold on $Sim_{WM}$. Other approaches store the normal similarity $Sim_M$ recorded when $M$ was last active. $M$ is then accepted if $Sim_{WM}$ is within some threshold $\epsilon$ of $Sim_M$.
The segment is considered a recurrence of the accepted $M$ with highest $Sim_{WM}$, or as a new concept if no $M$ was accepted.

A problem with this approach is that many meta-information features are not powerful enough to uniquely represent all concepts in a given data stream~\cite{hu2018detecting, ren2018knowledge}. For example two classifiers representing very different behaviour may achieve the same error rate on $W$.
Given a repository $R$ of stored concept representations $[M_i, \dots, M_n]$ constructed from some set of meta-information features $\mathit{MI}$, and a window $W$ drawn from one representation $M_a$, we define the discrimination ability of $\mathit{MI}$ as the mean difference between similarity $Sim_a$ testing concept representation $M_a$ on $W$, and similarities $Sim_i, \dots, Sim_n$ testing each representation $M_i \in R$ on $W$. 

Current approaches to concept drift detection and model selection, discussed in Section~\ref{Sec:PastWork}, often suffer from low discrimination ability.
This is common when only supervised or unsupervised meta-information features are used. 
 In Section~\ref{Sec:Evaluation}  we show that low discrimination ability reduces the performance of concept drift detection and model selection.

\section{FiCSUM Framework}\label{Sec:Method}
\begin{figure}
    \centering
    \includegraphics[width=0.5\textwidth]{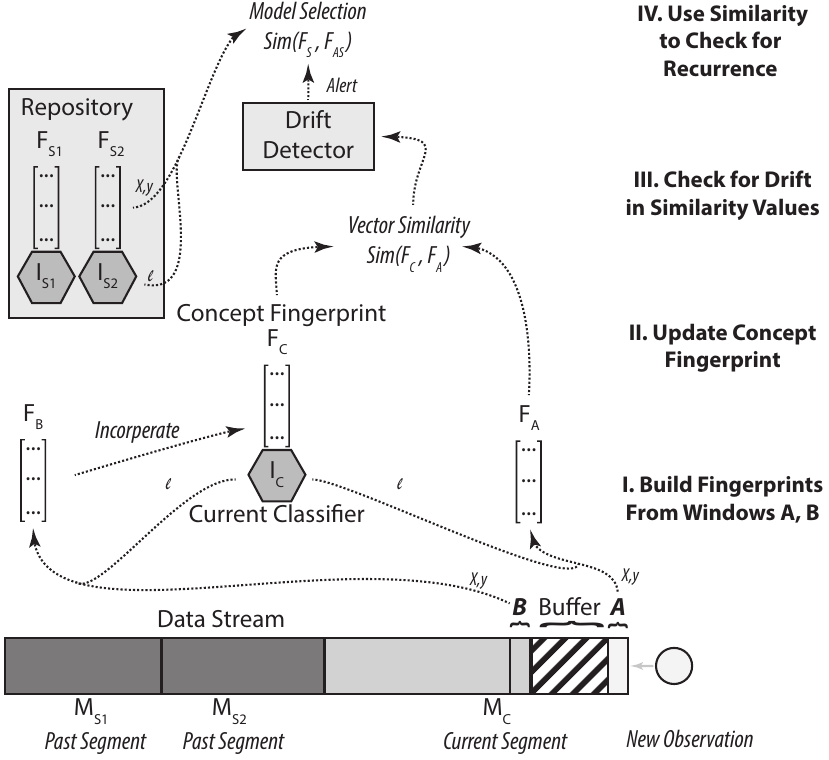}
    \caption{Concept Fingerprint Framework}
    \label{fig:diag}
\end{figure}
This section introduces our general framework, FiCSUM, to represent concepts using a combination of meta-information features.
FiCSUM represents concepts using a \textit{fingerprint}, a vector containing a diverse set of both supervised and unsupervised meta-information features. By combining many meta-information features, a fingerprint can uniquely identify a wider range of concepts than previous methods. This increases the performance of concept drift detection and model selection, and allows for more accurate tracking of stream concepts.
The set of meta-information features used in FiCSUM is general and flexible to support a wide range of datasets.
Similarly to selecting features for classification, in a given dataset some meta-information features may be irrelevant and introduce noise.
We propose a dynamic weighting strategy drawing on feature selection methods to learn which meta-information features can discriminate between concepts in a given dataset.

Figure~\ref{fig:diag} and Algorithm~\ref{Alg:FiCSUM} show the four basic steps in FiCSUM.
Sequential data stream observations are first collected into a window and transformed into a fingerprint representation as shown in Figure~\ref{fig:Fdiag}. 
Each element of a fingerprint is a meta-information feature distilling one aspect of concept behaviour.
A fingerprint constructed from a stationary window is used to update the \textit{concept fingerprint} representing the current stationary segment.
A fingerprint constructed from the most recent observations is tested against the concept fingerprint to check for concept drift.
If concept drift is detected, concept fingerprints stored in the repository are each tested for recurrence to determine the representation of the next stationary segment. 
This section describe these basic steps in more detail.

\subsection{Constructing and using Fingerprints}\label{sec:baseframework}

\begin{figure}
    \centering
    \includegraphics[width=0.5\textwidth]{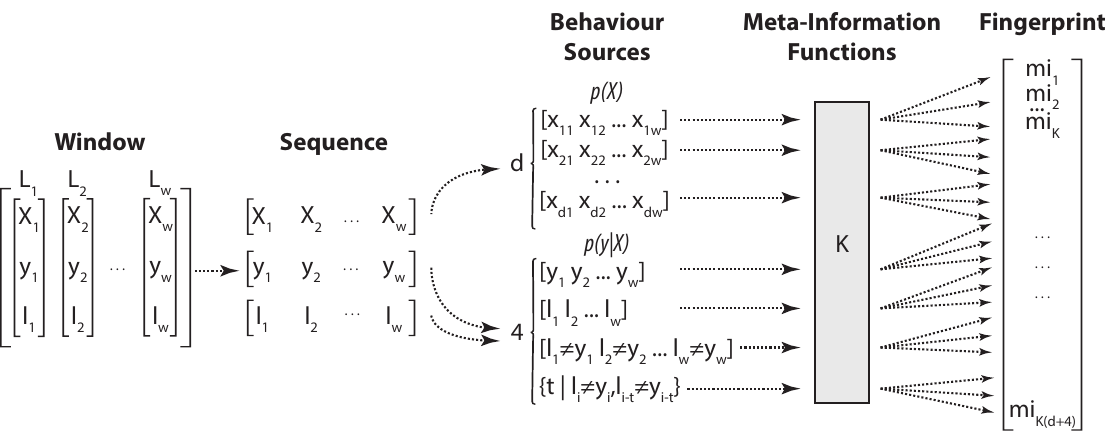}
    \caption{Constructing a fingerprint}
    \label{fig:Fdiag}
\end{figure}

\renewcommand{\algorithmicrequire}{\textbf{Input:}}
\renewcommand{\algorithmicensure}{\textbf{Output:}}
\begin{algorithm}
    \centering
    \begin{algorithmic}[1] 
        \Procedure{FiCSUM}{$DS$: a stream of observations, $w$: window size, $b$: buffer size, $P_C$: gap between fingerprint updates, $P_S$: gap between repository fingerprint updates}
            \State $F_c, I_c, \mu_c, \sigma_c \gets$ Initialize empty fingerprint, classifier and statistics
            \State $R \gets $ Initialize Repository
            \State $D \gets$ Initialize drift detector
            \State $Buf \gets []$ Initialize buffer
            \State $B, A \gets []$ Initialize buffer, active windows
            \While{$DS$ not empty, process observation $i$}
                \State $X, y \gets next(DS)$
                \State $l \gets I_c.$predict($X$)
                \State $I_c$.train($X$, $y$)
                \State $L \gets \langle X, y, l \rangle$
                \State $Buf$.append($L$), $A$.append($L$)
                \State $B$.append($Buf$.popElementOlderThan($b$))
                \State $B$.popElementOlderThan($w$)
                \State $A$.popElementOlderThan($w$)
                \If{$i \% P_C == 0$}
                \State $F_A \gets $makeFingerprint($A$) \Comment{Sec. ~\ref{sec:baseframework}}
                \State $F_B \gets $makeFingerprint($B$) \Comment{Sec. ~\ref{sec:baseframework}}
                \State $W \gets$ makeWeights($F_c$, $R$) \Comment{Sec. ~\ref{sec:weights}}
                \State $normSim \gets Sim(F_c, F_B, W)$ \Comment{Sec. ~\ref{sec:weights}}
                \State $\mu_c \gets$ onlineMean($\mu_c$, $normSim$)
                \State $\sigma_c \gets$ onlineSTDev($\sigma_c$, $normSim$)
                \State $F_c \gets $incorporate($F_c$, $F_B$)
                \State $D$.addToDetector($Sim(F_c, F_A, W)$)
                \If{$D$.alert()} \Comment{Model Selection}
                \State $R$.add($F_c, I_c, \mu_c, \sigma_c$)
                \State $accept \gets []$
                \For{$F_s$, $I_s$, $\mu_s$, $\sigma_s$ $\in R$}
                \State $F_{AS} \gets $makeFingerprint($I_s$.predict($A$))
                \If{$|Sim(F_{AS}, F_S, W) - \mu_s| \leq 2\sigma_s$}
                \State $accepted$.append($s$)
                \EndIf
                \EndFor
                \State $F_c, I_c, \mu_c, \sigma_c \gets \arg \max_{s} (accept)$ or New
                \EndIf
                \EndIf
                \If{$i\%P_S == 0$}
                \For{$F_s$, $I_s$, $\mu_s$, $\sigma_s$ $\in R$}
                \State $F_{SC} \gets $makeFingerprint($I_s$.predict($A$))
                \State $R$.updateNonActive($F_{SC}$) \Comment{Sec. ~\ref{sec:weights}}
                \EndFor
                \EndIf
            \EndWhile
        \EndProcedure
    \end{algorithmic}
    \caption{FiCSUM Algorithm}
    \label{Alg:FiCSUM}
\end{algorithm}
Figure~\ref{fig:Fdiag} describes how a window of $w$ observations, $W$, drawn from a concept $C$, as well as an incremental classifier trained on $C$, $I_C$, can be transformed into a vector representation of $C$, or \textit{concept fingerprint}.
The window is first separated into \textit{behaviour sources}, univariate sequences each describing a different aspect of behaviour.
The first $d$ behaviour sources are the sequences $x_{di} \in W$
describing the distribution $p(x_{d})$ for each input feature $x_d \in X$.
The next two behaviour sources, the sequences $y_i \in W$ and $l_i \in W$, describe the distributions  $p(y|X)$ shown by $C$ and $p(l|X)$ learned by $I_C$ respectively.
Two additional behaviour sources are constructed to describe the behaviour of classification errors. These are the sequence describing errors $l_i \ne y_i \in W$ and the sequence of distance between errors $\{t | l_i \ne y_i, l_{i-t} \ne y_{i-t}, \forall l_i,y_i \in W\}$.

Each behaviour source is distilled into a set of $K$ meta-information features through $K$ meta-information functions.
Each function maps a univariate sequence into a single value, for example mean or variance.
The observed range of each meta-information feature is scaled to the range $[0, 1]$.
Section~\ref{Sec:functions} further describes the meta-information functions used in FiCSUM.
Combining all meta-information features gives a $K(d+4)$ dimensional vector, or \textit{fingerprint}.
As an example, consider the window $[\langle X_1, y_1, l_1 \rangle, \langle X_2, y_2, l_2\rangle, \langle X_3, y_3, l_3\rangle]=[\langle[1,5],1,1\rangle,\langle[0.5,7],1,0\rangle,\langle[0.75,6],0,1\rangle]$. 
The behaviour sources $[1, 0.5, 0.75]$ and $[5, 7, 6]$ represent the first and second features in $X$. The sources $[1, 1, 0]$ and $[1, 0, 1]$ represent $y$ and $l$. Errors are represented as $[0, 1, 1]$, showing that $y_1 = l_1$, $y_2 \ne l_2$ and $y_3 \ne l_3$.
The gap between the two errors is 1, so the distance-between-errors behaviour source is $[1]$.
Using only the `mean' meta-information function, the fingerprint of the window would be: $[0.75, 6, 0.66, 0.66, 0.66, 1]$.

FiCSUM combines multiple fingerprints drawn from a single concept into a \textit{concept fingerprint}.
Each element of a concept fingerprint represents the observed distribution of a meta-information feature over the concept.
As FiCSUM is an online algorithm, this distribution is required to be calculated online in one pass, in constant time and space.
FiCSUM represents each meta-information feature as the triple $(\mu_{mi}, \sigma_{mi}, count_{mi})$ representing the mean, standard deviation and count across all incorporated fingerprints.
This specific representation is used to calculate the feature weights discussed in Section~\ref{sec:weights}.

Algorithm~\ref{Alg:FiCSUM} shows the overall flow of FiCSUM.
The first $w$ observations are used to initialize a concept fingerprint of the current concept, $F_c$, a classifier $I_c$ and an empty repository.
Periodically, FiCSUM captures an \textit{active window} $A$ as the $w$ most recent observations. If $A$ only contains observations from the current stationary segment, a fingerprint $F_A$ calculated from $A$ using classifier $I_c$ represents the current concept. If a concept drift has occurred, $F_A$ does not represent the current concept.
The similarity of $F_c$ to $F_A$, $Sim(F_c, F_A)$ measures likelihood that $A$ was drawn from the concept represented by $F_c$. The $Sim$ calculation is described in the next section.
A concept drift detector can identify when the similarity between $F_c$ and $F_A$ is significantly different to normal, \textit{i.e.}, a concept drift has occurred.
Here we use ADWIN to perform this concept drift detection.

The representation of the current concept, $F_c$, is trained by periodically incorporating fingerprints drawn from the current stationary segment.
These fingerprints cannot be constructed from $A$ as the most recent observations may be from a different distribution due to a delay in drift detection.
We make the assumption that any delay will be smaller than $b$ observations.
We can then assume that observations older that $b$ are drawn from the current concept if no drift has been detected.
FiCSUM collects a \textit{buffer window} $B$ of observations drawn only from the current concept by placing new observations in a buffer of length $b+w$ and selecting the oldest $w$.
A fingerprint $F_B$ constructed from the buffer window using classifier $I_c$ presents another representation of the current concept. 
The distribution of $Sim(F_c, F_B)$ is recorded as mean $\mu_c$ and standard deviation $\sigma_c$ in the repository to determine normal variation in stationary conditions.
$F_B$ is incorporated into $F_c$ by updating $\mu_{mi}$ and $\sigma_{mi}$ for each meta-information feature $mi$. The repository stores each $(F_c, I_c, \mu_c, \sigma_c)$ as a representation of concept $C$.

Model selection occurs when a concept drift is detected, to identify if recent observations $A$ are drawn from a concept $S$ stored in the repository.
For each stored concept representation, the stored classifier $I_S$ predicts new labels for $A$ to calculate fingerprint $F_{AS}$. A $Sim(F_{AS}, F_S)$ close to the normal operation for $S$ indicates that $A$ was drawn from $S$, and a recurrence is likely. We determine normal operation to be a similarity measure  of $\mu_S \pm 2\sigma_S$. 
The new segment is considered to be a recurrence of the maximum similarity stored concept meeting this criteria, setting $F_S$ and $I_S$ to be the active fingerprint and classifier.
If no stored concepts meet the criteria the segment is considered to be a new concept and a new fingerprint and classifier are initialized.

Often $A$ is partially drawn from the previous segment or concept drift region rather than fully drawn from the emerging segment.
If $A$ contains few observations from the emerging segment, $F_{AS}$ may not represent the emerging concept well.
A recurrence in this case is unlikely to be detected, creating a new concept representation.
 If $A$ is of size $w$ and is partially drawn from an emerging concept, then in at most $w$ observations $A$ will be fully drawn from the emerging concept.
FiCSUM performs a second model selection step $w$ observations after every concept drift. If a recurrence is found it is selected and the alternative is deleted.

\subsection{Similarity Calculation and Dynamic Weighting}\label{sec:weights}
A fingerprint is a vector representing system behaviour over a window of observations.
The $\mu_{mi}$ elements of a concept fingerprint can also be taken as a vector. 
Similarity between these vector representations can be calculated using a vector similarity measure.
Cosine distance is a popular measure of similarity between high dimensional vectors.
Given two n-dimensional fingerprint vectors $F_a$, $F_b$, cosine similarity $\mathit{Sim}(F_a, F_b)$ is calculated as:
$$
\mathit{Sim}(F_a, F_b) = \frac{F_a \cdot F_b}{||F_a|| \cdot ||F_b||}.
$$

Standard cosine similarity treats a difference in each dimension equally. However, here each dimension represents a different meta-information feature with its own range and discrimination ability. 
A deviation of the same magnitude in different meta-information features may have different impacts on fingerprint similarity.
For example, a difference in normalized error rate of 0.1 between fingerprints may be a strong indication of concept drift, while a difference in normalized skew of 0.1 may be standard in stationary conditions.
FiCSUM learns a vector of \textit{dynamic weights} $W$ to adjust the influence of each meta-information feature:
$$
Sim(F_a, F_b, W) = \frac{WF_a \cdot WF_b}{||WF_a|| \cdot ||WF_b||}.
$$
The weight $w_{mi} \in W$ of each meta-information feature $mi$ is learnt online per dataset.
Each $w_{mi}$ represents the scale of deviations normally seen in a feature, $w_{\sigma_{mi}}$, and the features discrimination ability, $w_{d_{mi}}$, calculated as:
$$
w_{mi} = w_{\sigma_{mi}} \times w_{d_{mi}}.
$$

The first weight, $w_{\sigma_{mi}}$, accounts for meta-information features operating at different scales.
Intuitively, the importance of a difference in meta-information values between fingerprints depends on previously seen differences. A difference of 0.1 in an $mi$ which normally has a standard deviation of 0.5 shows high similarity. 
However if normal standard deviation was 0.001 this difference would show low similarity.
Unweighted cosine distance does not take differences in scale into account.
The term $w_{\sigma_{mi}}$ scales each $mi$ such that differences are compared in terms of standard deviation, rather than raw magnitude. 
For each concept, FiCSUM stores the observed standard deviation of each meta-information feature $mi$ in the concept fingerprint as $\sigma_{mi}$, calculating $w_{\sigma_{mi}}$ as: 
$$
w_{\sigma_{mi}} = \frac{1}{\sigma_{mi}}.
$$

The second weight, $w_{d_{mi}}$, accounts for differences in discrimination ability. For a given dataset, changes in some meta-information features indicate concept drift while changes in others are noise.
Consider two concepts with error rate 0 and 1 respectively, and in both the mean of a given feature is normally distributed with a standard deviation of 0.5. Error rate and feature mean have equal variance but only error rate can differentiate the concepts. 
Here, error rate should be weighted higher to account for its increased importance to similarity.

Feature selection is a similar area which aims to select or weight input features ($X$) which provide more discrimination between class labels ($Y$)~\cite{li2017feature}.
Feature selection methods can be used to weight meta-information features which best discriminate between concepts.
We observe that there are two aspects of discriminating between concepts. A meta-information feature with high \textit{inter-concept} variation varies between a fingerprint calculated with classifier $I_A$ on observations drawn from concept $A$ and a fingerprint calculated with $I_B$ on observations drawn from $B$. This variation is important in distinguishing between stored fingerprints for model selection.
A feature with high \textit{intra-classifier} variation varies between fingerprints calculated with classifier $I_A$ on observations drawn from concept $A$ compared to observations drawn from concept $B$. This variation is important in detecting concept drift.

Feature selection methods can be used to assign weights based on these aspects.
Fisher score~\cite{li2017feature} is chosen here as it is efficient to calculate, and only requires a constant space to store the distribution of each feature as a mean and standard deviation.
Other methods \textit{e.g.} mutual information score, may be used but may require the distribution of each feature to be approximated in constant space.
FiCSUM calculates $w_{d_{mi}}$ as the maximum of $v_{s_{mi}}$ based on inter-concept variation and $v_{sc_{mi}}$ based on intra-classifier variation:
$$w_{d_{mi}} = \max(v_{s_{mi}}, v_{sc_{mi}}).$$
\subsubsection{Inter-concept variation}
Let $\mu^S_{mi}$, $\sigma^S_{mi}$ be the mean and standard deviation of meta-information feature $mi$ recorded by fingerprint $F_S\in R$ in repository $R$. The mean value of $mi$ across fingerprints is $\bar \mu_{mi} = E[\{\mu^S_{mi}|S\in R\}]$.
We calculate $v_{s_{mi}}$ using Fisher score, adjusted to compare the variation in $\mu^S_{mi}$ between concepts to the maximum $\sigma^S_{mi}$ within a concept. 
Using the maximum rather than mean $\sigma^S_{mi}$ avoids instability when concepts have zero or small variance.
$$
v_{s_{mi}} =\sqrt{\frac{\sum_S (\mu^S_{mi} - \bar \mu_{mi})^2}{|R|}} \times \frac{1}{\max(\{\sigma^S_{mi}| S\in R \})}.
$$

\subsubsection{Intra-classifier variation}
Given an active concept $C$ FiCSUM periodically captures a fingerprint $F_{SC}$ for each stored concept representation $S$ in the repository, representing the behaviour of $S$ on observations drawn from $C$.
We calculate $v_{sc_{mi}}$ using the Fisher score method as the variance between $F_S$ and $F_{SC}$ relative to variation in $F_S$ . 
Let $\mu^{SC}_{mi}$, $\sigma^{SC}_{mi}$ be the mean and standard deviation of $mi$ as recorded by fingerprint $F_{SC} | C \in R$. The mean of $\mu_{mi}$ across $F_S$ and $F_{SC}$ is $\bar \mu^{SC}_{mi} = E[\{ \mu^{SC}_{mi} | C \in R \}]$. Standard deviation across $F_C$ and $F_{SC}$ is then:
$$
 \sigma(\mu^{SC}_{mi}) = \sqrt{\frac{\sum_{C \in R}(\mu^{SC}_{mi} - \bar \mu^{SC}_{mi})^2}{|R|}}.
$$
The mean of $\sigma(\mu^{SC}_{mi})$ relative to $\sigma^{SC}_{mi}$ across all $S$ gives $v_{sc_{mi}}$:
$$
v_{sc_{mi}} = E\left[\left\{\frac{\sigma(\mu^{SC}_{mi})}{\sigma^{SC}_{mi}} \bigg | S \in R\right\}\right].
$$

\subsection{Meta-Information Features}\label{Sec:functions}
FiCSUM uses a set of meta-information features $\mathit{MI} = \{mi_i\}$ to differentiate between concepts $C = \{c_i\}$. 
Selecting \textit{MI} is similar to the area of feature selection, which selects a set of features to differentiate between classes. In this section, we discuss aspects of feature selection which can be applied to selecting \textit{MI}, \textit{relevancy}, \textit{coverage} and \textit{redundancy}~\cite{tang2014feature, li2017feature}.

Firstly, each $mi \in \mathit{MI}$ should be \textit{relevant}. In feature selection, relevance refers to correlation between a feature and class labels. Here we consider relevance to be correlation between $mi$ and $C$. 
Secondly, for all pairs of concepts $c_i, c_j \in C$ there should be some relevant  $ mi \in \mathit{MI}$ which can discriminate between $c_i$ and $c_j$.
We refer to an $\mathit{MI}$ containing at least one $mi$ capable of discriminating between any $c_i, c_j \in C$ as having high \textit{coverage}.
Lastly, each $mi \in \mathit{MI}$ should be \textit{non-redundant}. A pair of features $mi_i, mi_j \in \mathit{MI}$ is redundant if $mi_i$ is correlated with $mi_j$, or explains the same variation in concept.
Redundancy adds noise to the similarity calculation without adding discrimination ability.

As a guideline, we propose a set of meta-information features used to distinguish concepts have high \textit{coverage} and contain only \textit{relevant} meta-information features without \textit{redundancy}.
FiCSUM enforces relevancy online per dataset in the previously described dynamic weighting step. 
We enforce coverage and redundancy offline in the initial selection of \textit{MI}, selecting a large pool of non-correlated meta-information functions from a literature survey discussed in Section~\ref{Sec:PastWork}.

Selecting relevant features online from a diverse pool is common in feature selection~\cite{tang2014feature}.
However, this makes the assumption that features are independent, or non-redundant. 
An alternative in the batch domain is to calculate feature correlation on a training set, discarding any found to be correlated.
In the online case calculating feature correlation is not practical.
We instead assume meta-information correlation is \textit{static}, discarding features found to be correlated on in an offline evaluation on synthetic data.
Table~\ref{tab:mi_measures} shows the 13 meta-information functions selected for this work, and the behaviour sources used to calculate them.
In Section~\ref{Sec:Evaluation} we evaluate the individual performance of each meta-information function compared to the combined set used in FiCSUM. 

\begin{table}[]
    \caption{FiCSUM Meta-Information Functions and Behaviour Sources.}
    \label{tab:mi_measures}
    \centering
    \begin{adjustbox}{max width=0.5\textwidth}
        \begin{tabular}{ll}
            \multicolumn{1}{c}{Meta-Information Function} & \multicolumn{1}{c}{Behaviour} \\
            \midrule
            Mean~\cite{katakis2010tracking} & Distribution Center\\
            Standard Deviation~\cite{cavalcante2016fedd} & Distribution Variance\\
            Skew~\cite{cavalcante2016fedd}& Distribution Asymmetry\\
            Kurtosis~\cite{cavalcante2016fedd} & Distribution Tails\\
            Autocorrelation lag 1 \& 2~\cite{cavalcante2016fedd} & Temporal Dependence\\
            Partial Autocorrelation lag 1 \& 2~\cite{cavalcante2016fedd} & Temporal Dependence\\
            Mutual Information~\cite{cavalcante2016fedd} & Temporal Dependence\\
            Turning Point Rate~\cite{cavalcante2016fedd} & Rate of Oscillation \\
            Entropy of intrinsic mode functions 1 \& 2~\cite{ding2019entropy} & Behaviour across timescales \\
            Shapley Value~\cite{zheng2019labelless} & Feature Importance \\
        \bottomrule
        \rule{0pt}{1ex} & \\
            \multicolumn{1}{c}{Behaviour Source} & \multicolumn{1}{c}{Behaviour Type}\\
            \midrule
            Features & $p(X)$ \\
            labels & $p(y|X)$ \\
            Classifier Labels & Learned $p(y|X)$ \\
            Errors & Learned $p(y|X)$ \\
            Error Distances & Temporal $p(y|X)$ \\
        \bottomrule
        \end{tabular}
    \end{adjustbox}

\end{table}

\section{Implementation}
FiCSUM trains three components online, the classifier for each concept representation, the normalization of each meta-information feature and dynamic weights.
Training these components produces variation in similarity even during stationary segments.
This section describes two implementation measures to reduce sensitivity to this variation.

As the normalization and dynamic weighting components train, the similarity calculated between the same two fingerprints may change.
This is a problem when similarity values are recorded, such as $\mu_C$ and $\sigma_C$ stored in the repository, as these records may not be an accurate reflection of similarity in the future.
To avoid static similarity records becoming out of date, a small sample of fingerprint pairs are retained alongside recorded similarity values.
When records are accessed in the future, the new similarity value between these pairs can be calculated.
The ratio between the new and old similarities describes a transformation to convert similarity records calculated using an old weighting scheme into the new weighting scheme. 
This transform allows $\mu_C$ and $\sigma_C$ to be compared to similarity values calculated in the future.

The behaviour of the classifier associated with each concept will change significantly as it is trained. This change is rapid when the classifier is first created, and slows as it reaches a stable state. 
FiCSUM adapts concept representations to new behavior by incorporating new fingerprints.
However if behaviour changes faster than these updates dissimilarity may be observed even in a stationary segment.
FiCSUM identifies significant classifier training events and increases fingerprint plasticity to incorporate new behaviour faster. When a classifier has significantly changed, \textit{e.g.}, a decision tree has grown a new branch or the absolute change in weights in a neural network is above some threshold, the distribution of meta-information features associated with classifier behaviour, \textit{i.e.} depend on classifier labels $l_i$, are reset. This forgets old classifier behaviour, allowing new behaviour to be learned.
FiCSUM uses a tree based incremental classifier, Hoeffding Tree. Significant changes in classifier behaviour are identified when the tree learns a new branch.  

\section{Time and Memory Complexity}\label{Sec:Complexity}
Incrementally training a classifier and explicit concept drift detection is $O(1)$ per observation.
FiCSUM constructs a fingerprint periodically every $P_C$ observations, requiring a window of length $w$. Fingerprint construction time complexity depends on the meta-information functions chosen.
Here mutual information and empirical mode decomposition give a per observation time complexity of $O(\frac{1}{P_C}w \log w)$ to construct a fingerprint.
A non-active fingerprint is constructed for each concept in repository $R$ periodically every $P_S$ observations to calculate dynamic weights, with per observation time complexity $O(\frac{1}{P_S}|R|w\log w)$.
Model selection at each drift detection is also $O(|R|w\log w)$.
Given the probability of drift as $D$, this gives a per observation time complexity of $O(D|R|w\log w)$ for model selection. 
Overall time complexity is constant per observation, $O(\frac{1}{P_C}(w \log w) + (D + \frac{1}{P_S})(|R|w \log w))$. For comparison, an ensemble method runs $|R|$ classifiers at each observation, $O(|R|)$ or $O(|R| + D(|R|w \log w)$ if model selection is used. The next section shows the effects of parameters $P_C$, $P_S$ and $w$ on runtime.

FiCSUM stores a repository to handle recurring concepts. With classifier size $\gamma$, this requires $O(|R|\gamma)$ memory. FiCSUM additionally stores a fingerprint vector for each concept. Each fingerprint is size $fz(d+4)$ where $f$ is the number of meta-information functions used, $z$ is the size of the stored distribution of each meta-information feature and $d$ is the number of input features. Here, we use 13 meta-information functions and store each distribution as two floats, mean and standard deviation. A buffer of $b+w$ observations is also stored. FiCSUM requires $O((b+w) + fz(d+4)|R|\gamma)$ space.

\begin{figure}[t]
    \centering
    \includegraphics[width=0.42\textwidth]{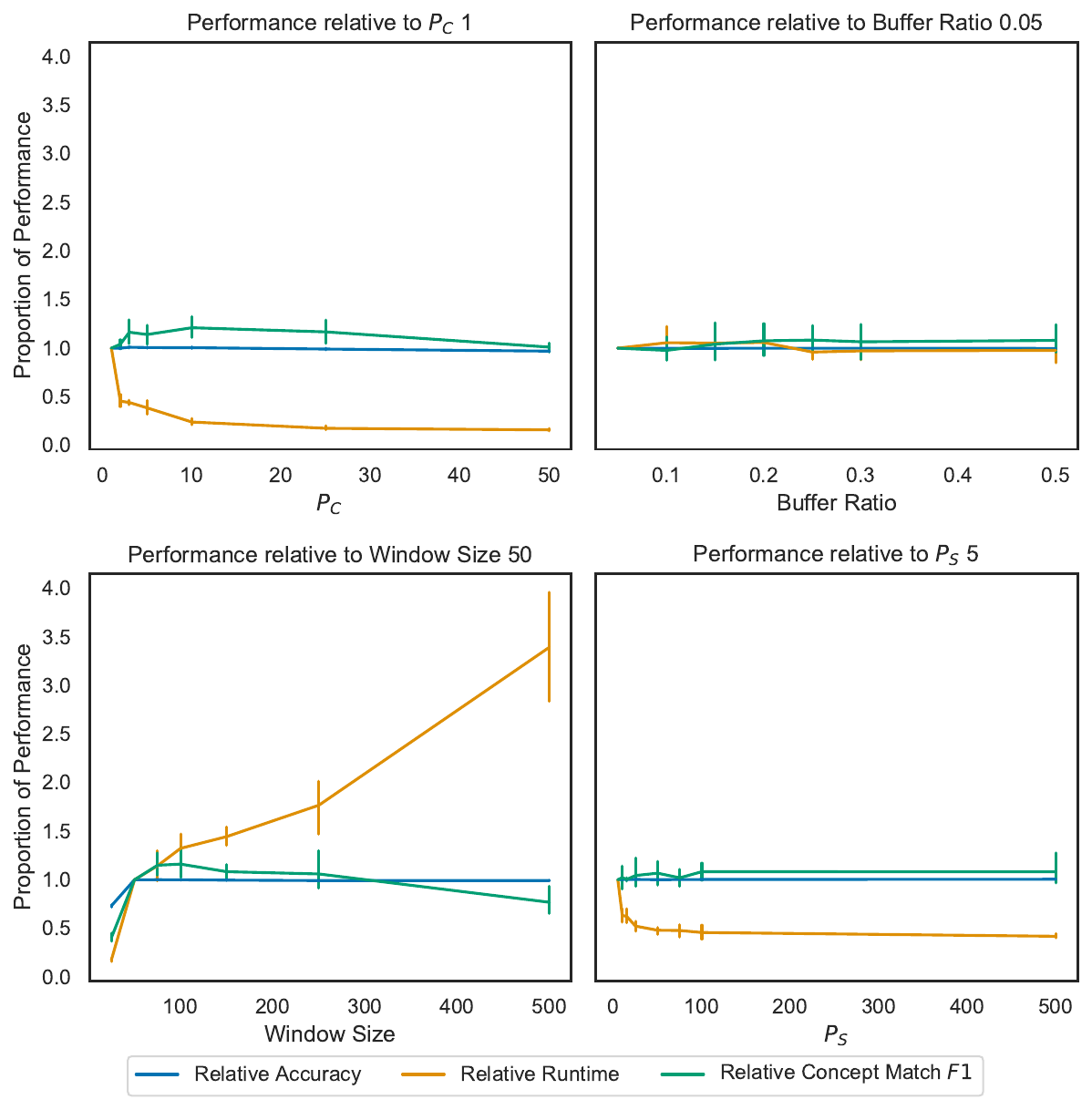}
    \caption{Sensitivity to changes in parameters in \textit{Arabic}. Values show the proportion of performance compared to a base parameter level.}
    \label{fig:perf_sensitivity}
\end{figure}
\begin{table}[b]
    \caption{Dataset Characteristics}
    \label{tab:datasets}
    \centering
    \begin{tabular}{lrrr}
    \multicolumn{1}{c}{Dataset} & \multicolumn{1}{c}{Length} & \multicolumn{1}{c}{\# features} & \multicolumn{1}{c}{\# contexts} \\
    \midrule
    AQTemp & 24000 & 25 & 6 \\
    AQSex & 24000 & 25 & 6 \\
    Arabic & 8800 & 10 & 10 \\
    CMC & 1473 & 8 & 2 \\
    QG & 4010 & 63 & 10 \\
    UCI-Wine & 6498 & 11 & 2\\
    \midrule
    RBF & 30000 & 10 & 6 \\
    RTREE & 30000 & 10 & 6 \\
    STAGGER & 30000 & 3 & 3\\
    HPLANE-U & 30000 & 10 & 6 \\
    RTREE-U & 30000 & 10 & 6\\
    \midrule
    \end{tabular}

\end{table}

\begin{table}[]
    \caption{Discrimination Ability of FiCSUM vs supervised (S-MI) and unsupervised (U-MI) meta-information, error rate (ER). Best of each dataset (row) bolded.}
    \label{tab:BDisc}
    \centering
    \begin{adjustbox}{max width=0.5\textwidth}

\begin{tabular}{lrrrr}
{} & \multicolumn{4}{c}{Discrimination Ability}\\
\multicolumn{1}{c}{DataSet} &            \multicolumn{1}{c}{ER} &         \multicolumn{1}{c}{S-MI} &         \multicolumn{1}{c}{U-MI} &         \multicolumn{1}{c}{FiCSUM} \\
\midrule
AQSex        &140.16 (199.20) &   173.15 (44.11) &    51.11 (23.19) &   \textbf{190.26 (44.98)} \\
AQTemp       &       8.83 (2.36) &   128.64 (83.31) &    71.15 (27.98) &   \textbf{184.91 (56.72)} \\
STAGGER     &   \textbf{963.32 (537.12)} &   339.1 (107.92) &      13.09 (3.20) &   138.55 (81.19) \\
RTREE        &         \textbf{6404.99 (8287.06)} &        87.73 (78.26) &    38.25 (15.42) &   289.15 (61.29) \\
RBF       &      10.29 (1.51) &   160.97 (49.69) &     22.75 (6.17) &   \textbf{224.33 (60.43)} \\
\midrule
Arabic       &     28.94 (12.16) &   106.24 (39.49) &  180.47 (117.21) &  \textbf{265.38 (144.11}) \\
CMC          &       1.12 (0.07) &    23.26 (22.81) &    20.25 (10.31) &    \textbf{60.64 (17.76)} \\
HPLANE-U &     18.31 (24.99) &   110.35 (22.34) &     74.01 (5.67) &  \textbf{215.56 (137.94}) \\
QG           &     18.43 (21.64) &     \textbf{90.53 (54.8)} &     25.78 (29.2) &     25.31 (24.5) \\
RTREE-U  &       8.81 (4.41) &   179.24 (62.37) &   129.96 (21.92) &  \textbf{ 222.17 (70.87)} \\
UCI-Wine     &       0.42 (0.01) &     45.5 (14.56) &    55.22 (43.63) &   \textbf{131.93 (11.41)} \\
\bottomrule
\end{tabular}

    \end{adjustbox}
\end{table}
\section{Evaluation}\label{Sec:Evaluation}
In this section, we evaluate the FiCSUM framework, investigating discrimination ability, performance, using the kappa statistic ($\kappa$), and ability to track concepts using the co-occurrence F1 (C-F1) measure discussed in Section~\ref{Sec:background}.
We first compared the performance of combining supervised and unsupervised meta-information features. We then compared the performance of each meta-information function on a range of datasets, showing the value of a diverse set. We finally compared to other state of the art frameworks.
\subsubsection{Datasets}
Evaluating the co-occurrence of concepts and representations requires datasets with known concepts.
We use six real-world datasets~\cite{moreira2018classifying} separated into distinct contexts, shown in Table~\ref{tab:datasets}.
 We compare to five synthetic datasets generated using the scikit-multiflow platform, each with six concepts.
 We note that in general, synthetic datasets display concept drift as a change in $p(y|X)$.
 In order to evaluate change in $p(X)$, in the HPLANE-U and RTREE-U datasets we induce change in $p(X)$ between concepts by setting the distribution, autocorrelation and frequency of the sampling function. 
 We note the effect of this change in the following sections.
 Each test was repeated 20 times using different concept and sampling seeds.
In order to create datasets with recurring concepts, we repeat each concept nine times, shuffling the order of appearance for each seed. For real world datasets, so as to not bias model selection each recurrence uses the next 75\% of concept observations. This ensures that a classifier trained on one occurrence has not seen the first 25\% of the next recurrence. Reducing data shared between recurrences would make each segment too small to learn a representation.

\subsubsection{Parameter Tuning}
FiCSUM features 4 parameters, window size $w$, buffer ratio, $P_C$ and $P_S$.
 We find that in general performance is not dependent on parameters as long as they are not set to extreme values, and that optimal parameter values are similar across datasets.
 We selected reasonable default parameters using a grid search on the \textit{Arabic} dataset, and validated performance on the RTREE dataset.
 Figure~\ref{fig:perf_sensitivity} shows the relative impact of changing each parameter in the \textit{Arabic} dataset.
 Accuracy rather than the $\kappa$ statistic is used to keep relative proportions above 0.
 Window size $w$ has the largest impact on performance.
 When $w$ is large relative to delay in concept drift detection, the window used for model selection is more likely to contain observations drawn from before the drift, reducing performance. On the other hand, meta-information features are calculated using $w$ and a larger $w$ may reduce variance. Calculating fingerprints on longer windows increases runtime as $w$ is increased. We set $w$ to 75, allowing for a small delay in concept drift detection and low runtime while also being large enough to avoid high variance in fingerprint calculation.
For both $P_C$ and $P_S$, there is a trade-off between performance and runtime. A low value is beneficial for performance but increases runtime. This trade-off can be tuned to suit the application. We set $P_C$ to 3 and $P_S$ to 25. For buffer ratio, a low value allows observations to be incorporated into our concept representation quicker, but with a higher chance of being drawn from the wrong concept. High values may be too conservative in incorporating new information producing a shallow drop off in performance. Performance peaks between 0.2 and 0.3 and there is no effect on runtime. We select a buffer ratio of 0.25.
Our implementation of FiCSUM uses the python based scikit-multiflow platform. We use the default Hoeffding Tree classifier and ADWIN drift detector implementations. All implementation and evaluation code, as well as additional experiments, are available at https://bit.ly/31008po.

\subsubsection{Behaviour Source Comparison}
In the following two experiments we validated the hypothesis that representing a concept using the full set of meta-information features outperforms using only supervised or unsupervised features.
We compare a variant of FiCSUM using all behaviour sources to variants using only unsupervised (U-MI) and supervised (S-MI) behaviour sources.
For the supervised case, we restrict behaviour sources to labels, predicted labels, errors and error distance. For the unsupervised case, we restrict behaviour sources to features. 
We also evaluate a variant restricted to the common error rate meta-information feature (ER).
\subsubsection{Discrimination ability}
We first compare discrimination ability as discussed in Section~\ref{sec:dd_ms}.
This measures the ability of each set of meta-information features to distinguish between concept representations.
It is important to note that the magnitude of discrimination ability does not directly affect performance, differentiating concepts only requires some threshold to be met.
Table~\ref{tab:BDisc} shows the discrimination ability of each system on each dataset. 
In the AQSex, AQTemp, CMC, STAGGER, RBF and RTREE datasets, unsupervised meta-information features have low discrimination ability. This indicates that concepts in these datasets are mainly differentiated by $p(y|X)$. STAGGER, RBF and RTREE are synthetic datasets where the main discriminant factor is change in labelling function. In Arabic and UCI-Wine, supervised features show low discrimination ability and
in CMC, RTREE\_U and UCI-Wine error rate displays very low discrimination ability, indicating concepts are differentiated by change in $p(X)$.

FiCSUM achieves the highest rank for discrimination ability in 8 of 11 datasets, indicating our fingerprint representation successfully captures the benefit of a general set of meta-information measures.
In the QG dataset, FiCSUM has a lower discrimination ability than both S-MI and U-MI. This was a short dataset with many possibly redundant features, which may have reduced the effectiveness of feature based meta-information.
In STAGGER and RTREE, only the labelling function changes between concepts. This means that accuracy along describes nearly all of the change, so it receives a very high discrimination ability. This also means that nearly all of the meta-information features contained in a fingerprint  are redundant. FiCSUM still receives a high discrimination ability relative to other methods. The following experiment shows that this allows FiCSUM to still distinguish between concepts.

\begin{table*}[]
    \caption{Performance using fingerprints for drift detection and model selection. Top segment shows datasets with mainly drift in $p(y|X)$, the bottom shows datasets with mainly drift in $p(X)$. Best of each dataset (row) bolded for each measure.}
    \label{tab:Neither}
    \centering
    \begin{adjustbox}{max width=\textwidth}
\begin{tabular}{lrrrr|rrrr}
{} & \multicolumn{4}{c}{$\kappa$ statistic} & \multicolumn{4}{c}{C-F1} \\
\multicolumn{1}{c}{Dataset} &              \multicolumn{1}{c}{ER} &           \multicolumn{1}{c}{S-MI} &           \multicolumn{1}{c}{U-MI} &           \multicolumn{1}{c}{FiCSUM} &               \multicolumn{1}{c}{ER} &          \multicolumn{1}{c}{S-MI} &           \multicolumn{1}{c}{U-MI} &           \multicolumn{1}{c}{FiCSUM} \\
\midrule
AQSex        &  0.93 (0.01) &   0.90 (0.02) &  0.71 (0.06) &  \textbf{0.94 (0.01)} &     0.51 (0.06) &  0.41 (0.04) &  0.65 (0.09) &  \textbf{0.75 (0.06)}\\
AQTemp       &  \textbf{0.58 (0.01)} &   0.50 (0.06) &  0.36 (0.05) &  0.47 (0.09) &     0.65 (0.05) &   0.49 (0.10) &  0.63 (0.08) &   \textbf{0.72 (0.10)} \\
STAGGER     &   \textbf{0.98 (0.00)} &  0.97 (0.01) &  0.41 (0.12) &  0.97 (0.02) &      \textbf{0.98 (0.00)} &  0.94 (0.04) &  0.48 (0.03) &  0.91 (0.06) \\
RBF       &  \textbf{0.75 (0.04)} &  0.72 (0.04) &  0.68 (0.03) &  0.73 (0.03) &      \textbf{0.82 (0.10)} &  0.67 (0.12) &  0.53 (0.11) &   0.73 (0.10) \\
RTREE        &  0.93 (0.07) &   0.79 (0.10) &  0.34 (0.08) &  \textbf{0.94 (0.04)} &     \textbf{0.76 (0.17)} &   0.50 (0.15) &   0.30 (0.04) &  0.74 (0.12) \\
\midrule
Arabic       &  \textbf{0.86 (0.02)} &  0.77 (0.06) &  0.85 (0.04) &  \textbf{0.86 (0.04)} &     0.57 (0.12) &  0.38 (0.05) &  \textbf{0.85 (0.06)} &  \textbf{0.85 (0.06)} \\
CMC          &  0.21 (0.02) &  0.22 (0.02) &  0.25 (0.03) &  \textbf{0.27 (0.02)} &     0.56 (0.05) &  0.61 (0.12) &   \textbf{0.80 (0.07)} &   0.76 (0.10)\\
HPLANE-U &  0.43 (0.02) &  0.42 (0.02) &  \textbf{0.44 (0.03)} &  \textbf{0.44 (0.03)} &     0.31 (0.03) &  0.28 (0.04) &  \textbf{0.95 (0.03)} &  0.75 (0.04) \\
QG           &   0.66 (0.10) &  0.59 (0.07) &  \textbf{0.73 (0.05)} &  0.72 (0.06) &     0.36 (0.05) &  0.32 (0.03) &  \textbf{0.52 (0.07)} &  \textbf{0.52 (0.06)}\\
RTREE-U  &  0.73 (0.07) &  0.68 (0.07) &  \textbf{0.81 (0.04)} &   0.80 (0.04) &     0.53 (0.09) &  0.47 (0.09) &  \textbf{0.95 (0.02)} &  0.91 (0.04) \\
UCI-Wine     &   0.20 (0.01) &   0.18 (0.00) &   \textbf{0.23 (0.00)} &  \textbf{0.23 (0.01)} &     0.54 (0.04) &  0.51 (0.01) &  0.73 (0.03) &  \textbf{0.92 (0.01)} \\
\midrule
Avg Rank & \textbf{2.17} & 3.11 & 2.83 & \textbf{1.89} & 2.49 & 3.43 & 2.32 & \textbf{1.76}\\
\bottomrule
\end{tabular}
\end{adjustbox}

\end{table*}

\subsubsection{Performance}
We evaluated the performance implications of the increased discrimination ability of FiCSUM against ER, S-MI and U-MI.
Table~\ref{tab:Neither} shows $\kappa$ statistic and C-F1 using each method.
The top segment of the table shows datasets where ER and S-MI achieved a $\kappa$ statistic more than two standard deviations higher than U-MI. 
In the bottom segment, U-MI achieved a co-occurrence F1 more than two standard deviations above ER and S-MI.
This indicates that unsupervised methods failed to identify concepts in five datasets and supervised methods failed to identify concepts in six datasets.
Generally supervised methods achieved a higher $\kappa$ statistic and C-F1 on the standard synthetic datasets STAGGER, RBF and RTREE. 
Unsupervised methods achieved a higher $\kappa$ statistic and C-F1 in datasets featuring large changes in feature distribution, such as in HPLANE-U and RTREE-U datasets where we injected change in $p(X)$.

FiCSUM displays high performance in both $\kappa$ statistic and C-F1 across all datasets, highlighting its generality.
The fingerprint representation avoids the substantially below optimal results on certain datasets seen in the baseline methods.
 FiCSUM achieved the highest $\kappa$ statistic performance in 6 of the 11 datasets and in the remainder is within two standard deviations of the best performer.
In five datasets FiCSUM achieves the highest co-occurrence F1 and in five is within two standard deviations of the best performing method.
We ranked each method across all datasets for both $\kappa$ statistic and C-F1, shown in the final row.
For both measures, a Friedman significance test rejects the null hypothesis that methods achieved the same rank with a $p$-value below 0.01.
The Nemenyi posthoc test at a significance level of $0.05$ finds that FiCSUM achieved a significantly higher $\kappa$ statistic than S-MI and U-MI and significantly better C-F1 than all three baselines.
This experiment was repeated isolating model selection by passing perfect drift detection signals and achieved similar results, shown in the linked supplementary material due to space constraints.
This validates our hypothesis that a combined set of meta-information features increases model selection and drift detection performance, and importantly, avoids the failure cases seen in the baseline methods.

\subsubsection{Meta-Information set comparison}
In this experiment, shown in Table~\ref{tab:MIFunc}, we investigated the relative discrimination ability of meta-information features across different types of drift induced in synthetic datasets.
 Outliers due to normalization are marked as $>500$.
We use the default random tree generator in scikit-multiflow and induce concept drift by changing the sampling of features in three ways per concept. We inject changes in feature distribution (mean, standard deviation, skew and kurtosis) in datasets labeled $Synth_D$, we inject feature autocorrelation in datasets labeled $Synth_A$ and we inject frequency in datasets labeled $Synth_F$ by overlaying a sine wave with amplitude and frequency per concept.

We find that the relative performance of each meta-information function varies between datasets. For example skew achieves a high $\kappa$ statistic relative to other functions when distributional change occur between concepts, but a relatively low $\kappa$ statistic when only autocorrelation or frequency changes. This validates our hypothesis that each meta-information measure is only able to represent some types of concept drift. 
We note that the combined set achieves the best or second best performance in all three measures across every dataset. This indicates that our dynamic weighting system successfully combines the best performing meta-information features for each dataset. 
\begin{table*}[]
    \caption{Meta-information functions with induced drift in distribution (D), autocorrelation (A) and frequency (F) of features. Best of each dataset (column) bolded for each measure.}
    \label{tab:MIFunc}
    \centering
    \begin{adjustbox}{max width=\textwidth}
\begin{tabular}{clrrrrrrr}
& &  \multicolumn{7}{c}{Dataset}\\
                                         & \multicolumn{1}{c}{Meta-Information Function} &        \multicolumn{1}{c}{$Synth_A$} &       \multicolumn{1}{c}{$Synth_{AF}$} &        \multicolumn{1}{c}{$Synth_D$} &                \multicolumn{1}{c}{$Synth_{DA}$} &      \multicolumn{1}{c}{$Synth_{DAF}$} &        \multicolumn{1}{c}{$Synth_{DF}$} &        \multicolumn{1}{c}{$Synth_{F}$} \\
\midrule
\multirow{11}{*}{\rotatebox[origin=c]{90}{$\kappa$ statistic}} & Shapley Value &         0.64 (0.05) &         0.53 (0.07) &         0.84 (0.04) &                0.88 (0.05) &         0.85 (0.03) &         0.79 (0.08) &        0.52 (0.08) \\
                                         & Mean &         0.96 (0.02) &         0.84 (0.04) &         0.97 (0.01) &                0.97 (0.01) &         0.97 (0.01) &         0.97 (0.02) &         \textbf{0.79 (0.10)} \\
                                         & Standard Deviation &         0.86 (0.09) &         0.44 (0.04) &          \textbf{0.98 (0.00)} &                \textbf{0.98 (0.01)} &         0.97 (0.01) &         0.97 (0.01) &         0.46 (0.10) \\
                                         & Skew &         0.51 (0.08) &          0.40 (0.05) &         0.92 (0.02) &                0.94 (0.01) &         0.91 (0.06) &         0.88 (0.02) &        0.52 (0.07) \\
                                         & Kurtosis &         0.59 (0.04) &         0.45 (0.07) &         0.94 (0.01) &                0.95 (0.01) &         0.94 (0.02) &          0.90 (0.03) &        0.41 (0.08) \\
                                         & Autocorrelation &         0.72 (0.08) &         0.51 (0.05) &         0.91 (0.02) &                0.93 (0.03) &         0.94 (0.02) &         0.92 (0.02) &        0.44 (0.08) \\
                                         & Partial Autocorrelation &         0.75 (0.13) &         0.62 (0.13) &         0.91 (0.03) &                0.95 (0.01) &         0.94 (0.02) &         0.89 (0.03) &        0.41 (0.08) \\
                                         & Mutual Information &          0.60 (0.09) &          0.44 (0.10) &          0.80 (0.05) &                0.89 (0.04) &         0.93 (0.02) &         0.87 (0.01) &        0.68 (0.09) \\
                                         & Turning point rate &         0.85 (0.12) &         0.83 (0.08) &         0.96 (0.02) &                0.96 (0.01) &         0.95 (0.02) &         0.94 (0.02) &        0.66 (0.15) \\
                                         & Entropy of intrinsic mode functions &          0.90 (0.04) &         0.49 (0.07) &         0.95 (0.02) &                0.95 (0.02) &         0.95 (0.03) &         0.92 (0.04) &        0.46 (0.16) \\
                                         & FiCSUM &          \textbf{0.97 (0.00)} &          \textbf{0.90 (0.06)} &         \textbf{0.98 (0.01)} &                \textbf{0.98 (0.01)} &         \textbf{0.98 (0.01)} &         \textbf{0.98 (0.01)} &        0.74 (0.11) \\
\midrule
\multirow{11}{*}{\rotatebox[origin=c]{90}{C-F1}} & Shapley Value &          0.40 (0.06) &         0.27 (0.03) &         0.48 (0.05) &                0.55 (0.07) &         0.43 (0.08) &         0.38 (0.11) &        0.25 (0.02) \\
                                         & Mean &         0.92 (0.05) &         0.61 (0.06) &         0.89 (0.09) &                 0.90 (0.06) &          0.87 (0.10) &         0.87 (0.09) &        0.37 (0.06) \\
                                         & Skew &         0.33 (0.04) &         0.29 (0.02) &         0.42 (0.06) &                0.47 (0.09) &           0.40 (0.10) &         0.35 (0.04) &        0.33 (0.02) \\
                                         & Standard Deviation &         0.79 (0.11) &         0.31 (0.02) &         \textbf{0.94 (0.03)} &                 \textbf{0.97 (0.00)} &          0.90 (0.03) &         0.92 (0.05) &        0.36 (0.08) \\
                                         & Kurtosis &         0.35 (0.04) &          0.30 (0.02) &         0.37 (0.09) &                0.39 (0.06) &         0.39 (0.07) &         0.38 (0.05) &        0.29 (0.02) \\
                                         & Autocorrelation &         0.64 (0.07) &         0.37 (0.07) &         0.34 (0.05) &                0.47 (0.06) &           0.60 (0.10) &         0.61 (0.07) &        0.29 (0.02) \\
                                         & Partial Autocorrelation &         0.63 (0.16) &          0.50 (0.08) &         0.38 (0.07) &                0.51 (0.09) &         0.49 (0.03) &         0.53 (0.08) &        0.27 (0.01) \\
                                         & Mutual Information &         0.43 (0.07) &         0.38 (0.05) &         0.27 (0.01) &                0.29 (0.01) &          0.51 (0.10) &         0.49 (0.04) &        \textbf{0.54 (0.11)} \\
                                         & Turning point rate &         0.75 (0.18) &         \textbf{0.78 (0.08)} &           0.70 (0.10) &                 0.66 (0.10) &          0.60 (0.05) &         0.72 (0.09) &         0.60 (0.12) \\
                                         & Entropy of intrinsic mode functions &         0.79 (0.07) &         0.35 (0.05) &         0.74 (0.07) &                0.67 (0.04) &           0.60 (0.10) &         0.65 (0.06) &        0.43 (0.11) \\
                                         & FiCSUM &         \textbf{0.96 (0.02)} &         0.69 (0.12) &         0.92 (0.04) &                0.93 (0.03) &         \textbf{0.91 (0.05)} &         \textbf{0.93 (0.03)} &        0.44 (0.09) \\
\midrule
\multirow{11}{*}{\rotatebox[origin=c]{90}{Discrimination}} & Shapley Value &       35.44 (32.18) &      \textbf{157.46 (54.89)} &     \textbf{\textgreater500 (\textgreater500)} &  \textbf{\textgreater500 (\textgreater500)} &     198.66 (127.17) &   \textbf{\textgreater500 (\textgreater500)} &      29.99 (21.88) \\
                                         & Mean &         86.20 (61.50) &         34.80 (18.50) &     274.82 (173.31) &            283.69 (160.28) &      190.84 (88.01) &      165.98 (24.09) &       39.22 (9.64) \\
                                         & Standard Deviation &       55.67 (22.59) &        12.26 (5.09) &       86.28 (26.33) &              99.77 (39.59) &         25.2 (9.98) &         29.06 (8.70) &        3.94 (1.32) \\
                                         & Skew &         3.86 (2.93) &         2.41 (0.31) &         9.48 (8.84) &                6.43 (2.87) &       11.07 (12.06) &         6.81 (5.26) &        2.62 (1.46) \\
                                         & Kurtosis &          2.40 (1.06) &         1.81 (0.45) &          6.20 (1.87) &                6.46 (1.88) &         5.46 (1.95) &         7.08 (4.27) &        3.19 (0.72) \\
                                         & Autocorrelation &        24.28 (6.82) &         10.6 (5.23) &         11.0 (3.71) &               20.92 (9.22) &        15.32 (5.82) &        20.21 (5.21) &      34.33 (10.27) \\
                                         & Partial Autocorrelation &        22.83 (8.58) &        12.76 (5.43) &        14.09 (4.08) &                16.46 (6.20) &        15.83 (7.25) &        15.73 (4.83) &        22.79 (9.90) \\
                                         & Mutual Information &           -$^*$ &           -$^*$ &           -$^*$ &                  -$^*$ &           -$^*$ &           -$^*$ &         50.00 (0.00) \\
                                         & Turning point rate &        13.74 (5.36) &        12.37 (3.11) &        26.86 (9.29) &               21.58 (8.08) &        23.25 (4.43) &       28.51 (12.84) &        4.91 (1.76) \\
                                         & Entropy of intrinsic mode functions &       51.06 (16.93) &       19.39 (15.08) &       55.23 (17.54) &               58.89 (9.57) &        51.12 (9.61) &       53.15 (22.32) &        5.72 (3.11) \\
                                         & FiCSUM &      \textbf{279.74 (56.36)} &       124.25 (33.20) &     416.29 (109.75) &             342.47 (86.49) &      \textbf{340.94 (76.53)} &      315.24 (49.09) &      \textbf{78.71 (33.96)} \\
\bottomrule
\end{tabular}
    \end{adjustbox}
    $^*$ Discrimination ability is displayed as \textgreater500 for values above 500 and $-$ for methods which only identified one concept. 
\end{table*}

\begin{table*}[]
    \caption{Performance Comparison Between Frameworks. Best of each dataset (column) bolded for each measure.}
    \label{tab:Framework}
    \centering
    \begin{adjustbox}{max width=\textwidth}

\begin{tabular}{llrrrrrrrrr}
             & \multicolumn{1}{c}{Framework} &                \multicolumn{1}{c}{AQSex} &              \multicolumn{1}{c}{CMC} &         \multicolumn{1}{c}{UCI-Wine} &             \multicolumn{1}{c}{RBF} &         \multicolumn{1}{c}{RTREE-U} &            \multicolumn{1}{c}{Arabic} &        \multicolumn{1}{c}{HPLANE-U} &                \multicolumn{1}{c}{QG} &            \multicolumn{1}{c}{STAGGER} \\
\midrule
\multirow{6}{*}{\rotatebox[origin=c]{90}{$\kappa$ statistic}} & HTCD &          0.94 (0.00) &      0.23 (0.05) &      0.21 (0.00) &        0.62 (0.02) &         0.57 (0.04) &       0.86 (0.01) &         0.42 (0.01) &       0.84 (0.01) &        0.95 (0.01) \\
             & RCD &          0.69 (0.01) &      0.17 (0.01) &      0.06 (0.00) &        0.52 (0.03) &         0.51 (0.07) &       0.74 (0.11) &         0.06 (0.05) &       0.54 (0.05) &        0.82 (0.04) \\
             & ER &          0.93 (0.01) &      0.20 (0.00) &      0.20 (0.01) &        0.79 (0.01) &         0.72 (0.04) &       0.81 (0.03) &         0.41 (0.01) &       0.59 (0.02) &        \textbf{0.99 (0.00)} \\
             & DWM &          0.88 (0.00) &      0.19 (0.00) &      0.18 (0.00) &        0.56 (0.02) &         0.49 (0.04) &       0.85 (0.03) &         0.42 (0.00) &       0.66 (0.03) &        0.91 (0.02) \\
             & ARF &          0.94 (0.00) &      \textbf{0.40 (0.01)} &      \textbf{0.34 (0.00)} &        \textbf{0.82 (0.01)} &         0.71 (0.03) &       \textbf{0.91 (0.04)} &         \textbf{0.48 (0.00)} &       \textbf{0.97 (0.01)} &        \textbf{0.99 (0.00)} \\
             & FiCSUM &          \textbf{0.95 (0.01)} &      0.30 (0.02) &      0.26 (0.01) &        0.81 (0.02) &         \textbf{0.83 (0.03)} &       0.90 (0.01) &         0.42 (0.02) &       0.84 (0.04) &        0.98 (0.01) \\
\midrule
\multirow{6}{*}{\rotatebox[origin=c]{90}{C-F1}} & HTCD &          0.12 (0.00) &      0.45 (0.14) &      0.13 (0.01) &        0.11 (0.00) &         0.11 (0.00) &       0.12 (0.01) &         0.18 (0.08) &       0.12 (0.00) &        0.11 (0.01) \\
             & RCD &          0.19 (0.02) &      0.45 (0.09) &      0.47 (0.00) &        0.29 (0.03) &         0.25 (0.03) &       0.27 (0.01) &         0.27 (0.01) &       0.28 (0.01) &        0.20 (0.00) \\
             & ER &          0.55 (0.01) &      0.62 (0.02) &      0.52 (0.00) &        0.84 (0.05) &         0.53 (0.06) &       0.45 (0.06) &         0.34 (0.02) &       0.34 (0.05) &        \textbf{0.98 (0.00)} \\
             & DWM &          0.29 (0.00) &      0.67 (0.00) &      0.63 (0.00) &        0.29 (0.00) &         0.29 (0.00) &       0.29 (0.00) &         0.29 (0.00) &       0.29 (0.00) &        0.50 (0.00) \\
             & ARF &          0.29 (0.00) &      0.67 (0.00) &      0.63 (0.00) &        0.29 (0.00) &         0.29 (0.00) &       0.29 (0.00) &         0.29 (0.00) &       0.29 (0.00) &        0.50 (0.00) \\
             & FiCSUM &          \textbf{0.80 (0.07)} &      \textbf{0.80 (0.06)} &      \textbf{0.71 (0.09)} &        \textbf{0.88 (0.09)} &         \textbf{0.94 (0.04)} &       \textbf{0.83 (0.08)} &         \textbf{0.78 (0.03)} &       \textbf{0.64 (0.04)} &        0.96 (0.02) \\
\midrule
\multirow{6}{*}{\rotatebox[origin=c]{90}{Runtime (s)}} & HTCD &        581.31 (4.88) &     21.07 (0.15) &    147.34 (0.83) &     \textbf{700.51 (16.20)} &       449.62 (1.38) &     151.12 (1.40) &      508.45 (12.97) &     133.71 (2.60) &      321.01 (4.51) \\
             & RCD &  41393.30 (11853.64) &  402.13 (171.27) &   1323.12 (9.69) &  9225.21 (2850.85) &  11493.21 (1023.98) &  2328.68 (535.42) &    8220.22 (930.11) &  2581.04 (568.57) &  10198.77 (127.11) \\
             & Acc &       605.04 (55.66) &     17.40 (1.50) &   123.93 (11.40) &   2651.41 (411.91) &      485.77 (46.04) &    175.25 (14.43) &      477.18 (54.14) &    112.62 (23.98) &     270.69 (21.68) \\
             & DWM &       \textbf{518.90 (39.79)} &     \textbf{13.64 (0.23)} &    \textbf{140.28 (0.79)} &     961.08 (10.77) &       \textbf{274.04 (0.83)} &     \textbf{86.61 (22.55)} &       \textbf{336.43 (4.02)} &      \textbf{82.01 (3.08)} &     \textbf{133.46 (12.41)} \\
             & ARF &      1582.78 (60.44) &     90.58 (1.87) &   812.35 (17.49) &   4163.04 (102.18) &     1700.22 (41.16) &    451.60 (15.50) &     2208.43 (32.37) &    257.88 (10.86) &     957.12 (21.57) \\
             & FiCSUM &     8171.87 (867.39) &   137.70 (18.50) &  1061.45 (66.56) &   5640.01 (786.48) &    6642.76 (886.49) &  1567.25 (196.60) &  11928.28 (2037.25) &   1198.45 (70.93) &  3581.32 (1027.54) \\
 \bottomrule
\end{tabular}

    \end{adjustbox}

\end{table*}
\subsubsection{Alternative frameworks}
 Table~\ref{tab:Framework}  compares the performance of FiCSUM to other frameworks: HTCD, a baseline Hoeffding Tree classifier reset when drift in error rate is detected using ADWIN, RCD, a recurring concept framework, two state of the art ensemble methods DWM and ARF and ER, a variant of our system using only error rate.
We use the MOA implementation of RCD with a Hoeffding Tree classifier and EDDM detector. We use the default DWM and ARF implementations in scikit-multiflow with 10 trees.

FiCSUM achieved the highest $\kappa$ statistic in the AQSex and RTREE-U datasets, and  ARF achieved the highest in the remaining seven datasets. 
Ensemble systems can be expected to achieve lower error rates than single classifier systems~\cite{anderson2016cpf}.
However, ensemble systems use a single evolving concept representation and do not track recurring concepts.
We achieved substantially better C-F1 results in tracking concepts than all comparison methods in all datasets apart from STAGGER.

FiCSUM's runtime depends on the meta-information features chosen. Table~\ref{tab:Framework} shows FiCSUM is generally slower than other methods, driven by the mutual information and EMD calculation. As shown in Figure~\ref{fig:perf_sensitivity} the selection of the $P_C$, $P_S$ and $w$ can be easily tuned to substantially reduce runtime.

\section{Related Work}\label{Sec:PastWork}
In this section, we discuss previously proposed concept representations for drift detection or model selection.

Supervised meta-information features, commonly error rate, can distinguish differences in $p(y|X)$ between concepts.
Error rate can be expected to decrease in stationary conditions and increase in non-stationary conditions~\cite{gonccalves2013rcd}.
Many drift detection methods such as ADWIN~\cite{bifet2007learning}, DDM~\cite{gama2004learning} and HDDM~\cite{frias2014online} are designed around this principle, detecting change in error rate as concept drift. 
Similarly, EDDM~\cite{baena2006early} uses the distance between errors to detect drift.
AUC has also been used to detect concept drift~\cite{wang2020auc}.
Supervised meta-information is used for concept drift detection in the recurrent concept systems RCD~\cite{gonccalves2013rcd}, CPF~\cite{anderson2016cpf}, JIT~\cite{alippi2013just} and DiversityPool~\cite{chiu2018diversity}.
For model selection, DiversityPool records the error rate of each stored classifier. A stored classifier achieving an error rate on a testing window within a 95\% confidence interval of its recorded error rate is considered a recurrence.
CPF and JIT store observations as a representation, and test the proportion of equivalent predictions made by a new classifier.

Other meta-information features depend only on classifier behaviour.
L-CODE~\cite{zheng2019labelless} uses change in feature importance to detect drift.
 Margin Density Drift Detection~\cite{sethi2017reliable} detects drift as a change in the proportion of observations close to a decision boundary. Both methods can be considered classifier specific, with fast Shapley value and uncertainty calculation available to tree and SVM based classifiers respectively.

Unsupervised meta-information features distinguish differences in $p(X)$. These changes often occur alongside changes in $p(y|X)$, so can often identify real concept drifts.
RCD represents concepts as a window of observations, which are tested against using a statistical distribution test. Comparing the entire distribution is slow, so other methods use meta-information features as a proxy for feature distribution.
JIT compares the mean and variance of each feature for both concept drift detection and model selection. CCP~\cite{katakis2010tracking} refers to this approach as a conceptual vector, and uses euclidean distance to calculate $Sim$. GraphPool~\cite{ahmadi2018modeling} extends this approach by also including feature correlation and using a likelihood measure to calculate $Sim$. Neither approach considers supervised meta-information.
The Fourier transform~\cite{da2017multidimensional} and empirical mode decomposition~\cite{ding2019entropy} methods allow the frequency of features to be used as meta-information features, and have been shown to detect concept drift.
Kuncheva et al.~\cite{kuncheva2013pca} apply PCA to observations and take the lowest variance components as meta-information features, showing that they are sensitive to concept drift.
In FEDD~\cite{cavalcante2016fedd} a range of meta-information features represent a concept, including variance, skew, kurtosis, turning point rate, correlation, partial correlation and mutual information, which is tested against using cosine similarity. This is similar to concept fingerprint, however it is limited to storing and testing the current concept representation for drift and does not consider supervised features.

Recent work recognizes that combining supervised and unsupervised features can detect more types of concept drift.
EFDD~\cite{hu2018detecting} proposes an ensemble of drift detectors, based on margin density, clustering and grid based measures.
In KME~\cite{ren2018knowledge} an ensemble of drift detectors based on accuracy, feature mean and variance is used.
In both cases, each meta-information feature is handled separately with individual drift detection schemes and similarity measures. 
A voting scheme combines each feature into a single result.
Extending either system requires a new drift detection scheme and similarity measure for each new meta-information measure.
In contrast, FiCSUM can handle adding meta-information measures without any additional architectural changes.

\section{Conclusion}
We propose the FiCSUM framework to represent the behaviour of data stream concepts as a fingerprint vector, made up of supervised and unsupervised meta-information features.
A novel dynamic weighting scheme learns meta-information influence online per dataset, allowing FiCSUM to be used on a general range of datasets, avoiding the failure cases seen in alternative methods.
Evaluation over 11 real and synthetic datasets verifies that the fingerprint representation allows a system to achieve significantly higher performance and tracking of ground truth concepts than representations using only supervised or unsupervised methods.
FiCSUM outperforms 4 adaptive frameworks including a state-of-the-art ensemble in both classification accuracy and modeling stream concepts.
FiCSUM allows increased flexibility in selecting meta-information features.
Future work utilizing this flexibility to dynamically adjust the set of meta-information features online could improve performance, and allow FiCSUM to adapt to periods of missing or delayed labels.

\section*{Acknowledgment}
The work was supported by the Marsden Fund Council from New Zealand Government funding (Project ID 18-UOA-005), managed by Royal Society Te Ap{\=a}rangi.


\bibliographystyle{IEEEtran}
\bibliography{IEEEabrv, refs.bib}
\end{document}